\documentclass[conference]{IEEEtran}
\IEEEoverridecommandlockouts
\usepackage{cite}
\usepackage{amsmath,amssymb,amsfonts}
\usepackage{algorithmic}
\usepackage{graphicx}
\usepackage{textcomp}
\usepackage{xcolor}
\usepackage{url}
\usepackage[hidelinks]{hyperref}

\def\BibTeX{{\rm B\kern-.05em{\sc i\kern-.025em b}\kern-.08em
    T\kern-.1667em\lower.7ex\hbox{E}\kern-.125emX}}
\begin{document}

\title{Uncertainty-Aware Post-Detection Framework for Enhanced Fire and Smoke Detection in Compact Deep Learning Models
{\footnotesize \textsuperscript{*}\\Accepted at the \href{https://smartmultimedia.org/2025/}{International Conference on Smart Multimedia (ICSM 2025)}}
\thanks{© 2025 IEEE. Personal use of this material is permitted. Permission from IEEE must be obtained for all other uses, 
including reprinting/republishing for advertising or promotional purposes, creating new collective works, 
for resale or redistribution to servers or lists, or reuse of any copyrighted component of this work in other works.}
}

\author{\IEEEauthorblockN{1\textsuperscript{st} Aniruddha Srinivas Joshi}
\IEEEauthorblockA{\textit{Independent Researcher} \\
aniruddha980@gmail.com}
\and
\IEEEauthorblockN{2\textsuperscript{nd} Godwyn James William}
\IEEEauthorblockA{\textit{Independent Researcher} \\
jamesgodwynj@gmail.com}
\and
\IEEEauthorblockN{3\textsuperscript{rd} Shreyas Srinivas Joshi}
\IEEEauthorblockA{\textit{Independent Researcher} \\
shreyassj2009@gmail.com}
}

\maketitle

\begin{abstract}
Accurate fire and smoke detection is critical for safety and disaster response, yet existing vision-based methods face challenges in balancing efficiency and reliability. Compact deep learning models such as YOLOv5n and YOLOv8n are widely adopted for deployment on UAVs, CCTV systems, and IoT devices, but their reduced capacity often results in false positives and missed detections. Conventional post-detection methods such as Non-Maximum Suppression and Soft-NMS rely only on spatial overlap, which can suppress true positives or retain false alarms in cluttered or ambiguous fire scenes. To address these limitations, we propose an uncertainty aware post-detection framework that rescales detection confidences using both statistical uncertainty and domain relevant visual cues. A lightweight Confidence Refinement Network integrates uncertainty estimates with color, edge, and texture features to adjust detection scores without modifying the base model. Experiments on the D-Fire dataset demonstrate improved precision, recall, and mean average precision compared to existing baselines, with only modest computational overhead. These results highlight the effectiveness of post-detection rescoring in enhancing the robustness of compact deep learning models for real-world fire and smoke detection.
\end{abstract}


\section{Introduction}

Accurate fire and smoke detection is vital for safety, surveillance, and disaster response, where early warnings can prevent severe damage and loss of life. Traditional sensor-based methods remain common but suffer from range and deployment limitations \cite{joglar2005probabilistic}. Vision-based approaches have emerged as strong alternatives, leveraging image analysis techniques such as color and motion cues \cite{chen2004early, yu2013real}.

Despite progress, two challenges persist: (i) compact detectors such as YOLOv5n and YOLOv8n are efficient for edge and IoT deployment but sacrifice accuracy, and (ii) standard post-detection methods like Non-Maximum Suppression (NMS) and its variants often suppress true positives or retain false positives in crowded or ambiguous scenes \cite{bodla2017softnms}. This creates a gap between speed and reliability in real-world fire detection.

We address this gap with a post-detection confidence rescoring framework that refines detector outputs using uncertainty estimates and visual plausibility features. A lightweight Confidence Rescoring Network (CRN) combines these cues to adjust box confidences, improving robustness without altering the detector. Our work investigates three research questions:
\begin{itemize}
    \item \textbf{RQ1:} Does rescoring improve detection performance over existing post-detection techniques when applied to compact detectors on fire/smoke imagery?
    \item \textbf{RQ2:} What is the relative contribution of uncertainty and visual features to performance gains?
    \item \textbf{RQ3:} What is the trade-off between accuracy improvements and computational overhead?
\end{itemize}

\section{Related Works}

\subsection{Compact Models for Fire and Smoke Detection}

Compact neural networks enable real-time detection in constrained environments. MobileNet \cite{howard2017mobilenets}, FireNet \cite{jadon2019firenet}, and generic detectors such as YOLO \cite{redmon2016yolo} along with their compact variants achieve fast inference but generally trade accuracy for speed and size compared to larger backbones. Earlier image-based methods relied on handcrafted cues such as fire-colored regions \cite{chen2004early, celik2007fire} or motion dynamics \cite{yu2013real}, but these approaches can be sensitive to lighting changes and scene clutter.

Compact models are particularly attractive in fire and smoke detection because deployment often occurs on UAVs, CCTV surveillance systems, or IoT devices with limited compute \cite{howard2017mobilenets, jadon2019firenet}. Studies show that CNN-based methods achieve higher accuracy than traditional handcrafted approaches, though early designs were computationally heavy \cite{muhammad2018early}. Subsequent work has moved toward lightweight or pruned networks that maintain accuracy while reducing memory and latency requirements, making them suitable for embedded fire detection \cite{devenancio2022}. In practice, compact detectors are often the realistic choice for field deployment, whereas heavier backbones are less feasible due to hardware or energy constraints. However, this efficiency may reduce representational capacity, potentially affecting recall and precision compared to larger backbones \cite{howard2017mobilenets}.

Recent work highlights increasing feasibility of compact detectors in safety-critical monitoring tasks. Their small memory footprint allows deployment on single-board computers and low-power hardware \cite{jadon2019firenet, devenancio2022}, yet this very efficiency can amplify vulnerability to ambiguous visual patterns. In such conditions, domain-specific post-detection strategies become especially important, as the base model has limited capacity to learn fine-grained distinctions on its own.

\subsection{Post-Detection Methods for Object Detection}

Post-detection methods play a crucial role in filtering detections. Classical Non-Maximum Suppression (NMS) \cite{neubeck2006efficient} and its variant Soft-NMS \cite{bodla2017softnms} remain widely adopted due to simplicity and speed. However, both operate solely on bounding-box overlap (IoU), which can be problematic in crowded or overlapping scenarios such as flames or smoke plumes. In such cases, true detections may be suppressed or false alarms retained.

Learned strategies such as Learned-NMS \cite{hosang2017learning} move beyond heuristic suppression by training models to refine detections jointly. These approaches achieve higher accuracy in generic object detection, yet they require additional training and introduce architectural complexity, which can be a concern in resource-constrained deployments. Furthermore, existing learned refinements are mostly developed for datasets such as COCO \cite{lin2014coco} and PETS \cite{ferryman2010pets}, with limited attention to domain-specific cases like fire and smoke.

This motivates the need for a post-detection approach that is lightweight, domain-aware, and specifically designed to enhance compact models without imposing significant computational costs.

\subsection{Challenges in Fire and Smoke Detection}

Prior fire and smoke detection studies have advanced along two largely separate tracks: lightweight architectures designed for efficient deployment \cite{jadon2019firenet, howard2017mobilenets, devenancio2022, muhammad2018early}, and handcrafted cues based on color or motion \cite{chen2004early, celik2007fire, yu2013real}. Both approaches show promise but can be sensitive to cluttered scenes, complex lighting, and overlapping fire and smoke regions.

Generic detectors such as YOLO \cite{redmon2016yolo} are often adopted, yet most pipelines continue to rely on standard NMS or Soft-NMS \cite{bodla2017softnms, neubeck2006efficient}, even though these methods may suppress true detections or retain false alarms in ambiguous cases. Adaptive refinements like Learned-NMS \cite{hosang2017learning} have been developed in generic detection, but they have not been applied to fire and smoke detection, where false negatives carry high risks.

\subsection{Broader Context in Fire and Smoke Detection}

Beyond vision-based approaches, probabilistic and sensor-driven fire detection methods have also been proposed. For instance, Joglar et al. \cite{joglar2005probabilistic} developed a probabilistic extension of the DETACT model to estimate activation times of ceiling-mounted detectors by incorporating parameter uncertainty. Such methods are valuable for system-level risk and reliability analysis but do not address the visual detection challenges posed in complex real-world imagery.

\subsection{Gap and Motivation}

Taken together, three gaps emerge. First, compact detectors are attractive for deployment on low-power devices but remain constrained by fixed post-detection heuristics. Second, learned refinements exist in general vision tasks but are not represented in the fire and smoke literature. Third, the integration of uncertainty with domain-relevant features for rescoring detections remains largely unexplored in the fire and smoke detection literature. Addressing this intersection motivates our work, which introduces a post-detection framework tailored to compact deep learning models for fire and smoke detection.

\section{Proposed Method}

In this section, we present our post-detection rescoring framework for improving fire and smoke detection in static images. The proposed framework refines the confidence scores of detections from a base object detection model (e.g., YOLO~\cite{redmon2016yolo}) by incorporating model uncertainty estimation and detection region feature analysis. Unlike traditional post-detection methods such as Non-Maximum Suppression (NMS), our framework introduces a learned Confidence Refinement Network (CRN) to replace heuristic-based confidence adjustments, resulting in a more adaptive and robust detection pipeline.

\begin{figure}[t]
    \centering
    \includegraphics[width=0.8\linewidth, height=15cm, keepaspectratio]{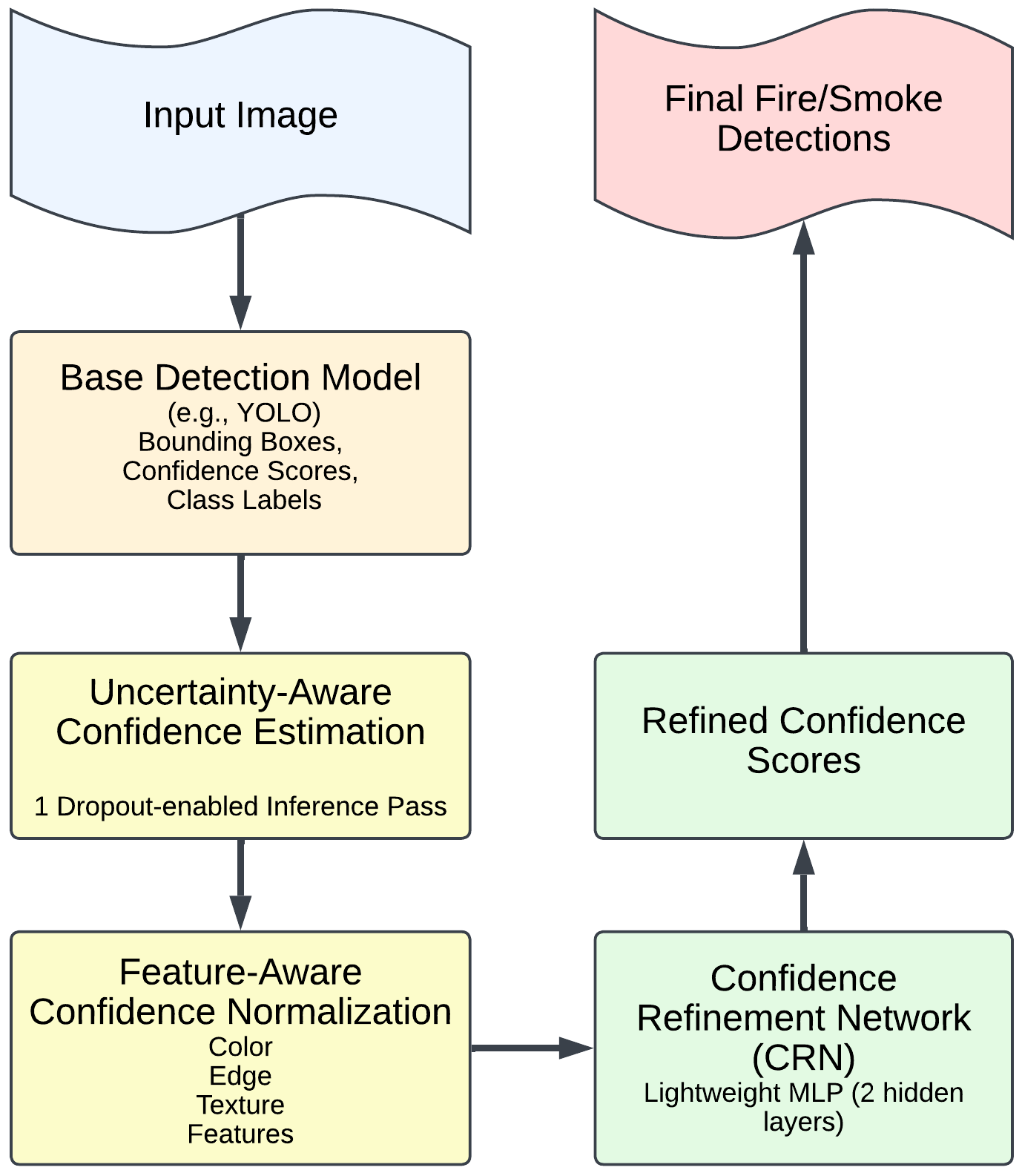}
    \caption{Pipeline of the proposed framework. The framework refines confidence scores using uncertainty estimation, detection region features, and a learned CRN.}
    \label{fig:pipeline}
\end{figure}

The proposed framework operates as a model-agnostic post-detection step, applicable to any object detection model, improving reliability without modifying the base detector. It consists of three key components which are illustrated in Fig~\ref{fig:pipeline}.

\subsection{Uncertainty-Aware Confidence Estimation}

To account for model uncertainty, we enable dropout during inference in the base detection model. Dropout, typically used during training, is repurposed here to estimate confidence variability. Unlike Monte Carlo Dropout, which requires multiple inference passes~\cite{gal2016dropout}, we adopt a single-pass approximation to maintain computational efficiency while capturing sufficient uncertainty information. Specifically, a single forward pass is performed with dropout activated, and the confidence scores of the detected bounding boxes are used to compute the mean confidence $\mu_c$ and variance $\sigma_c^2$:

\begin{equation}
    \mu_c = c_i
\end{equation}

\begin{equation}
    \sigma_c^2 = \frac{1}{N} \sum_{j=1}^{N} (c_j - \mu_c)^2
\end{equation}

where $N$ is the number of inference passes (default = 5 in training, reduced to 1 for deployment), $c_j$ is the confidence score from inference pass $j$, and $\mu_c$ is the mean confidence score across passes.

Dropout, when applied during inference, can be interpreted as a Bayesian approximation, where the stochasticity introduced by dropout provides a distribution over model predictions~\cite{gal2016dropout}. This allows for the estimation of model uncertainty without requiring multiple forward passes, as the dropout-induced randomness approximates the variability observed in Bayesian methods. While multi-pass Monte Carlo Dropout provides a more exhaustive uncertainty estimate, recent studies have shown that single-pass dropout can still capture meaningful uncertainty patterns in deep networks with limited computational overhead~\cite{kendall2017uncertainties}. This makes single-pass dropout particularly suitable for real-time applications where computational efficiency is critical. Empirical results (see Section~\ref{sec:performance}) confirm that this single-pass approximation sufficiently refines confidence scores while maintaining computational efficiency.

\subsection{Feature-Aware Confidence Normalization}

To further refine confidence scores, we incorporate detection region feature analysis. Each detected bounding box is evaluated based on its color, edge, and texture characteristics, ensuring that the predicted fire and smoke regions align with expected visual properties. While this study focuses on fire and smoke detection, the feature-based confidence refinement approach is adaptable to other object categories by selecting domain-relevant visual features.

The selected features are based on domain-specific characteristics of fire and smoke regions. Color intensity and saturation are extracted using HSV histograms, as fire regions exhibit strong saturation in the red-orange spectrum, while smoke regions appear more diffuse. HSV color space is widely used in fire detection due to its ability to separate chromatic and intensity information, making it more effective than RGB for distinguishing fire and smoke. This is consistent with the physical properties of fire and smoke, where fire emits bright, saturated colors due to its combustion process, while smoke appears less saturated as it consists of dispersed particles that scatter light, reducing color intensity.

Edge consistency is computed using Canny edge detection, as false positives often exhibit sharp, unnatural edges, whereas actual fire and smoke regions demonstrate gradient-like diffusion. The Canny algorithm is optimal for detecting edges while suppressing noise, making it effective for capturing the smooth transitions in fire and smoke boundaries~\cite{canny1986edge}. This aligns with the observation that fire and smoke produce smoother gradients due to their dynamic and fluid nature.

Texture randomness is captured using Haralick texture features, specifically contrast and homogeneity, which quantify high-frequency patterns typical of fire and the smoother textures of smoke~\cite{haralick1973texture}. These features have been widely adopted in fire and smoke classification tasks due to their ability to differentiate structured and diffuse textures. These distinctions are crucial for differentiating between fire/smoke and other textured objects in the scene.

These features are normalized and combined with the raw confidence score and uncertainty measure to form a feature vector $\mathbf{f}$:

\begin{equation}
    \mathbf{f} = [c_i, \sigma_c^2, s_i, e_i, t_i]
\end{equation}

where $s_i$, $e_i$, and $t_i$ represent the color, edge, and texture features, respectively.

\subsection{Confidence Refinement Network (CRN)}

At the core of our framework lies the Confidence Refinement Network (CRN), a compact feedforward neural network designed for efficiency trained to optimize confidence scores based on statistical uncertainty and feature-based plausibility. The CRN architecture comprises four stages:
\begin{enumerate}
    
\item Input Layer: The input to the CRN is a 5-dimensional feature vector $\mathbf{f} = [c_i, \sigma_c^2, s_i, e_i, t_i]$, which includes the raw confidence score, uncertainty estimate, and detection region features. This feature vector is designed to capture both the statistical uncertainty and visual characteristics of the detection.

\item Fully Connected Layer 2: The first hidden layer consists of 32 neurons with ReLU activation. This choice aligns with prior studies on network initialization and efficient layer sizing, which emphasize balancing model complexity and computational efficiency~\cite{glorot2010understanding}. ReLU activation is chosen for its ability to mitigate the vanishing gradient problem and accelerate convergence~\cite{nair2010rectified}. The output of this layer is computed as:
   \begin{equation}
       \mathbf{h}_1 = \text{ReLU}(\mathbf{W}_1 \mathbf{f} + \mathbf{b}_1)
   \end{equation}

\item Fully Connected Layer 2: The second hidden layer also consists of 32 neurons with ReLU activation. The output of this layer is computed as:
   \begin{equation}
       \mathbf{h}_2 = \text{ReLU}(\mathbf{W}_2 \mathbf{h}_1 + \mathbf{b}_2)
   \end{equation}

\item Output Layer: The final layer consists of a single neuron with sigmoid activation, which outputs the refined confidence score $\hat{c}_i$:
   \begin{equation}
       \hat{c}_i = \sigma(\mathbf{W}_3 \mathbf{h}_2 + \mathbf{b}_3)
   \end{equation}
\end{enumerate}

Here, $\mathbf{W}_1, \mathbf{W}_2, \mathbf{W}_3$ are weight matrices, $\mathbf{b}_1, \mathbf{b}_2, \mathbf{b}_3$ are bias vectors, and $\sigma$ is the sigmoid activation function, ensuring that confidence scores remain in the range $[0,1]$.

\subsection{Final Decision Processing}

Refined confidence scores are thresholded at $\tau$ (set to 0.5) to remove uncertain detections. A threshold of 0.5 is a commonly used practice in object detection frameworks such as YOLO~\cite{redmon2016yolo} to filter low-confidence predictions. Predictions with lower confidence scores are often associated with higher uncertainty and a higher likelihood of being false positives, as object detectors tend to be overwhelmed by easy negatives, which contribute to misclassifications~\cite{lin2017focal}. While adaptive thresholds can be dataset-dependent, a fixed threshold of 0.5 has been widely adopted for balancing precision and recall in object detection models. By discarding such detections, we ensure that only high-confidence predictions are retained, improving the overall reliability of the detection pipeline.

Since the CRN dynamically refines confidence scores, many false-positive overlapping detections are naturally suppressed through confidence thresholding. This is because overlapping detections often receive lower confidence scores due to their ambiguous nature. While Soft-NMS~\cite{bodla2017softnms} provides an alternative approach that retains overlapping detections by adjusting confidence scores, our method achieves suppression through feature-driven confidence refinement rather than spatial overlap considerations. Traditional NMS can still be applied externally if additional filtering is needed.

\begin{figure}[t]
    \centering
    \includegraphics[width=1.0\linewidth, height=10cm, keepaspectratio]{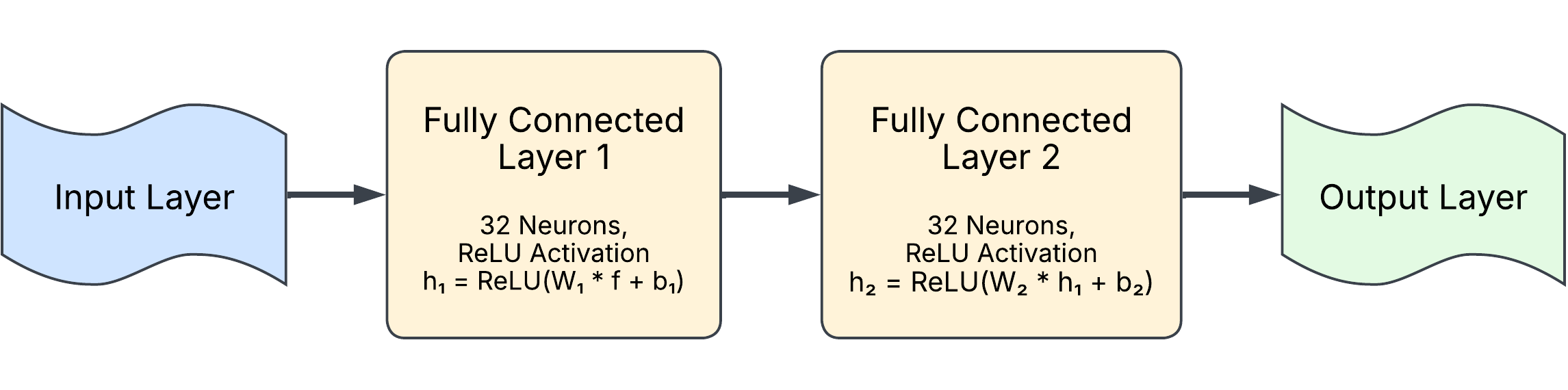}
    \caption{Architecture of the Confidence Refinement Network (CRN). The network refines confidence scores using uncertainty estimates and visual features.}
    \label{fig:crn}
\end{figure}

\section{Experimental Setup}
\label{section_experiment_setup}

We evaluated our framework using YOLOv5n and YOLOv8n from their official repositories \cite{yolov5repo, yolov8repo}, implemented in PyTorch 1.13 \cite{paszke2019pytorch} with CUDA 12.1 and cuDNN 8.9.2. Experiments were conducted on a workstation with an AMD Ryzen 9 5900X CPU, 64GB RAM, and an NVIDIA RTX 4070 Super GPU.

\subsection{Dataset}

The D-Fire dataset \cite{devenancio2022} was used for evaluating our approach. This benchmark dataset contains 21,527 images annotated with fire and smoke bounding boxes. In total, the dataset includes 14,692 bounding boxes for fire instances and 11,865 bounding boxes for smoke instances. Dataset details are in Table~\ref{table_dataset}.

\begin{table}[h]
\centering
\caption{Dataset Breakdown}
\label{table_dataset}
\begin{tabular}{|l|c|}
\hline
\textbf{Category} & \textbf{Number of Images} \\ \hline
Only Fire & 1,164 \\ \hline
Only Smoke & 5,867 \\ \hline
Fire and Smoke & 4,658 \\ \hline
None (Negative Samples) & 9,838 \\ \hline
\end{tabular}
\end{table}


The D-Fire dataset is particularly suitable for evaluating compact fire and smoke detection models, as prior work \cite{devenancio2022} has shown its applicability on embedded systems using Tiny YOLOv4 \cite{bochkovskiy2020yolov4}.

\subsection{Model Training and Configuration}

We trained two compact YOLO models, YOLOv8n and YOLOv5n, using the D-Fire dataset. These models were chosen for their efficiency in real-time detection tasks. YOLOv5n has 1.9M parameters, a model size of 3.8MB, and an inference time of 0.2ms per image on an RTX 4070 Super. YOLOv8n has 3.2M parameters, a model size of 5.5MB, and an inference time of 0.18ms per image on the same hardware. The Confidence Refinement Network (CRN) was trained using predictions generated by these models. Training was conducted until convergence using the Adam optimizer \cite{kingma2017adammethodstochasticoptimization} with a learning rate of 0.001 and a binary cross-entropy loss function.

\subsection{Post-Detection Techniques}

We benchmarked our framework against several established post-detection techniques: (i) Non-Maximum Suppression (NMS) \cite{neubeck2006efficient}, which removes overlapping boxes above IoU 0.45, (ii) Soft-NMS \cite{bodla2017softnms}, which decays scores using a Gaussian function ($\sigma = 0.5$), (iii) Edge-Based Filtering (EBF) \cite{canny1986edge}, which discards boxes with sharp edges assuming fire and smoke exhibit smooth gradients, (iv) Color-Based Filtering (CBF), which applies RGB thresholds ($R>200$, $G>100$, $B<100$ for fire; $R<100$, $G<100$, $B>200$ for smoke), (v) Histogram-Based Color Filtering (HBCF), which leverages HSV distributions, and (vi) Spatial Context Filtering (SCF), which requires fire and smoke detections to co-occur spatially.

\section{Performance Analysis and Comparison}\label{sec:performance}
\begin{figure*}[t]
    \centering
    \includegraphics[width=\textwidth]{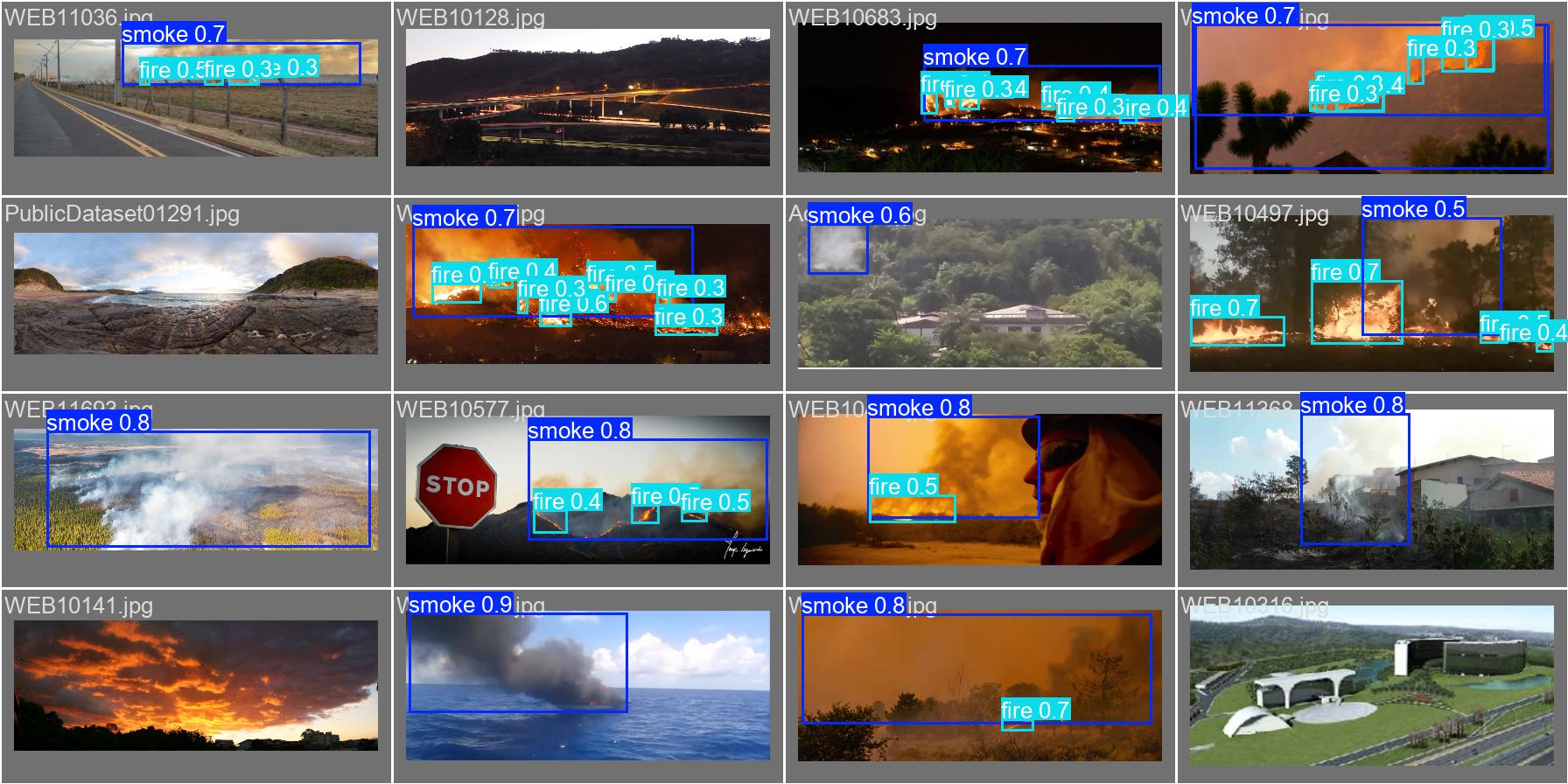}
    \caption{Qualitative results produced by the proposed method. The examples illustrate the effectiveness of the approach across different scenarios.}
    \label{fig:results}
\end{figure*}

We evaluated our proposed framework using the setup described in Section~\ref{section_experiment_setup}. Results reported in Tables~\ref{tab:yolov8_results} and~\ref{tab:yolov5_results}, and qualitative examples are shown in Figure~\ref{fig:results}.

\begin{table*}[t]
\centering
\caption{Performance comparison of YOLOv8n with different post-detection techniques.}
\label{tab:yolov8_results}
\begin{tabular}{|l|c|c|c|c|c|c|c|c|}
\hline
\textbf{Metric} & \textbf{YOLOv8n} & \textbf{NMS} & \textbf{Soft NMS} & \textbf{EBF} & \textbf{CBF} & \textbf{HBCF} & \textbf{SCF} & \textbf{Proposed Method} \\ \hline
Precision @ IOU = 0.5 & 0.712 & 0.763 & 0.712 & 0.773 & 0.788 & 0.762 & 0.781 & 0.845 \\ \hline
Recall @ IOU = 0.5 & 0.674 & 0.669 & 0.701 & 0.624 & 0.535 & 0.395 & 0.551 & 0.82 \\ \hline
mAP@50 & 0.625 & 0.627 & 0.657 & 0.592 & 0.499 & 0.377 & 0.553 & 0.651 \\ \hline
mAP@50-95 & 0.363 & 0.36 & 0.383 & 0.344 & 0.293 & 0.212 & 0.313 & 0.4 \\ \hline
Smoke Precision @ IOU = 0.5 & 0.769 & 0.814 & 0.769 & 0.817 & 0.858 & 0.815 & 0.815 & 0.86 \\ \hline
Smoke Recall @ IOU = 0.5 & 0.769 & 0.764 & 0.831 & 0.759 & 0.532 & 0.541 & 0.762 & 0.84 \\ \hline
Smoke AP@IOU = 0.5 & 0.72 & 0.719 & 0.777 & 0.715 & 0.507 & 0.511 & 0.718 & 0.78 \\ \hline
Fire Precision @ IOU = 0.5 & 0.661 & 0.717 & 0.661 & 0.728 & 0.74 & 0.692 & 0.731 & 0.82 \\ \hline
Fire Recall @ IOU = 0.5 & 0.598 & 0.592 & 0.604 & 0.516 & 0.537 & 0.278 & 0.382 & 0.79 \\ \hline
Fire AP@ IOU = 0.5 & 0.531 & 0.535 & 0.536 & 0.47 & 0.49 & 0.243 & 0.348 & 0.6 \\ \hline
Avg End-to-End Proc. Time (ms) & 12.78 & 13.31 & 15.81 & 14.9 & 15.56 & 15.96 & 16.63 & 20.15 \\ \hline
\end{tabular}
\end{table*}

\begin{table*}[t]
\centering
\caption{Performance comparison of YOLOv5n with different post-detection techniques.}
\label{tab:yolov5_results}
\begin{tabular}{|l|c|c|c|c|c|c|c|c|}
\hline
\textbf{Metric} & \textbf{YOLOv5n} & \textbf{NMS} & \textbf{Soft NMS} & \textbf{EBF} & \textbf{CBF} & \textbf{HBCF} & \textbf{SCF} & \textbf{Proposed Method} \\ \hline
Precision @ IOU = 0.5 & 0.703 & 0.762 & 0.764 & 0.775 & 0.754 & 0.758 & 0.776 & 0.84 \\ \hline
Recall @ IOU = 0.5 & 0.659 & 0.65 & 0.648 & 0.651 & 0.328 & 0.379 & 0.728 & 0.818 \\ \hline
mAP@50 & 0.609 & 0.609 & 0.607 & 0.579 & 0.298 & 0.36 & 0.511 & 0.641 \\ \hline
mAP@50-95 & 0.35 & 0.346 & 0.345 & 0.333 & 0.163 & 0.2 & 0.298 & 0.37 \\ \hline
Smoke Precision @ IOU = 0.5 & 0.757 & 0.819 & 0.82 & 0.823 & 0.801 & 0.825 & 0.82 & 0.85 \\ \hline
Smoke Recall @ IOU = 0.5 & 0.748 & 0.74 & 0.739 & 0.736 & 0.286 & 0.519 & 0.739 & 0.83 \\ \hline
Smoke AP@IOU = 0.5 & 0.7 & 0.7 & 0.699 & 0.696 & 0.27 & 0.492 & 0.699 & 0.68 \\ \hline
Fire Precision @ IOU = 0.5 & 0.655 & 0.711 & 0.714 & 0.726 & 0.727 & 0.673 & 0.713 & 0.82 \\ \hline
Fire Recall @ IOU = 0.5 & 0.588 & 0.577 & 0.575 & 0.509 & 0.361 & 0.267 & 0.359 & 0.78 \\ \hline
Fire AP@ IOU = 0.5 & 0.517 & 0.517 & 0.516 & 0.461 & 0.325 & 0.228 & 0.324 & 0.62 \\ \hline
Avg End-to-End Proc. Time (ms) & 14.23 & 14.59 & 14.28 & 16.47 & 16.61 & 16.89 & 17.7 & 23.148 \\ \hline
\end{tabular}
\end{table*}

\subsection{Quantitative Results for YOLOv8n}

\begin{itemize}
    \item \textit{Precision/Recall:} Improved from 0.712/0.674 to 0.845/0.82, showing a better balance of true positives and fewer false alarms than all baselines.
    \item \textit{mAP:} Increased from 0.625 to 0.651, with stronger mAP@50--95 (0.40 vs. 0.383 for Soft-NMS), indicating better localization consistency.
    \item \textit{Class-wise:} Fire AP rose from 0.531 to 0.60 and Smoke AP from 0.72 to 0.78, highlighting robustness across both categories.
\end{itemize}

\subsection{Quantitative Results for YOLOv5n}
\begin{itemize}
    \item \textit{Precision/Recall:} Improved from 0.703/0.659 to 0.84/0.818, with the main benefit coming from reducing false positives and negatives in fire cases.
    \item \textit{mAP:} Raised from 0.609 to 0.641, the most substantial gain among tested methods, confirming the effectiveness of rescoring.
    \item \textit{Class-wise:} Fire AP increased from 0.517 to 0.62, while Smoke AP remained stable (0.70 to 0.68), suggesting the framework’s strongest impact was on fire detection.
\end{itemize}

\section{Discussion}
Our experimental results validate the effectiveness of our uncertainty-aware post-detection framework. In this section, we contextualize these findings by addressing the research questions we posed earlier, discussing the broader implications of our work, and acknowledging the limitations of our study.

\subsection{Answering Our Research Questions}
\noindent\textbf{RQ1: Does rescoring improve detection performance over existing post-detection techniques when applied to compact detectors on fire/smoke imagery?}\\
Yes, our rescoring framework significantly improves detection performance. Our results consistently show that our method surpasses standard post-detection techniques like NMS, Soft-NMS, and various feature-based filtering methods across both YOLOv8n and YOLOv5n models. Our framework achieves a better trade-off between precision and recall, leading to higher overall reliability. Unlike NMS, which only considers spatial overlap, our approach integrates model uncertainty and visual plausibility, allowing it to suppress false positives and recover true positives that heuristic methods might miss.

\medskip
\noindent\textbf{RQ2: What is the relative contribution of uncertainty and visual features to performance gains?}\\
Our proposed framework's strength lies in the synergistic combination of statistical uncertainty and domain-relevant visual features. Uncertainty Estimation, derived from single-pass dropout inference, provides insight into the model's confidence variability. This helps identify detections where the model is hesitant, which often correspond to ambiguous or challenging scenes. Visual Features (color, edge, and texture) ground the detections in physical reality. By analyzing HSV histograms for color saturation, Canny edges for gradient smoothness, and Haralick features for texture randomness, our Confidence Refinement Network (CRN) learns to penalize detections that do not align with the typical appearance of fire or smoke.


\medskip
\noindent\textbf{RQ3: What is the trade-off between accuracy improvements and computational overhead?}\\
We found a clear and justifiable trade-off. Our framework boosts precision and recall by over 10 absolute percentage points in most cases, a level of improvement that is critical for safety applications where false negatives or false positives carry high risks. At the same time, the end-to-end processing time increases modestly compared to baseline models and simpler post-detection steps, rising from 12.78 ms to 20.15 ms for YOLOv8n and from 14.23 ms to 23.148 ms for YOLOv5n. This additional latency is primarily due to feature extraction and the CRN forward pass, but it remains well within the requirements for many real-time deployments, particularly fixed surveillance systems (CCTV) where sub-second responses are acceptable. Overall, this balance between modest latency and substantial accuracy gains is particularly valuable for compact detectors, enabling practical deployment without relying on heavier, more power-intensive models.

\subsection{Implications and Limitations}
Our work addresses a critical gap in fire and smoke detection research: improving the reliability of compact models post-detection. Our framework offers a model-agnostic solution that can be integrated with existing compact detectors to enhance their performance without retraining or altering the base architecture. This is particularly relevant for deployment on edge devices where computational resources are limited.

However, our study has some limitations. First, our uncertainty estimation relies on a single-pass dropout approximation, which we chose to maintain efficiency, though it is less exhaustive than multi-pass Monte Carlo methods. Second, we handcrafted and optimized the visual features for fire and smoke; adapting our framework to other object detection domains would require selecting new, domain-relevant features. Finally, we conducted our evaluation on the D-Fire dataset, and performance may vary on datasets with different environmental conditions.

\subsection{Future Work}
Future work can build on our framework in three directions. First, while we adopted a single-pass dropout scheme for efficiency, alternative uncertainty estimation strategies could be explored to better quantify confidence without sacrificing real-time performance. Second, extending the framework from static images to video-based detection would allow the use of temporal information such as motion cues and frame-to-frame consistency, which are highly relevant to the dynamic nature of fire and smoke. Finally, we aim to validate the adaptability of our approach in other domains where compact detectors struggle with reliability, requiring the design of new domain-specific visual features and evaluations on diverse benchmark datasets. Together, these directions can broaden the applicability of our framework while maintaining its suitability for resource-constrained deployments.

\section{Conclusion}
In this paper, we introduced an uncertainty-aware post-detection framework designed to enhance the reliability of compact deep learning models for fire and smoke detection. By integrating statistical uncertainty with domain-specific visual features, such as color, edge, and texture, our lightweight confidence refinement network (CRN) rescales detection confidences to better reflect their true plausibility.

Our experiments, which we conducted on the D-Fire dataset with YOLOv5n and YOLOv8n models, demonstrate that this approach yields significant improvements in precision, recall, and mean average precision compared to standard post-detection methods such as NMS and Soft-NMS. Our framework successfully addresses the inherent trade-off between the efficiency of compact models and the accuracy required for critical safety applications, doing so with a modest and manageable increase in computational overhead. By refining detections rather than simply filtering them, our work provides a practical and effective pathway to building more robust vision-based fire safety systems for real-world deployment.

\end{document}